\documentclass[10pt,twocolumn,letterpaper]{article}

\usepackage{iccv}
\usepackage{times}
\usepackage{epsfig}
\usepackage{graphicx}
\usepackage{amsmath}
\usepackage{amssymb}

\usepackage{multicol}
\usepackage{multirow}
\usepackage[ruled,vlined]{algorithm2e}
\usepackage{subfigure}
\usepackage{booktabs}
\usepackage{array}
\usepackage{colortbl}
\usepackage[symbol]{footmisc}
\usepackage[pagebackref=true,breaklinks=true,letterpaper=true,colorlinks,bookmarks=false]{hyperref}

\iccvfinalcopy 

\def\iccvPaperID{} 

\ificcvfinal\fi

\definecolor{mygray}{gray}{.9}
\newcommand{\proposed}{{FIA}}

\begin{document}

\title{Feature Importance-aware Transferable Adversarial Attacks}


\author{Zhibo Wang$^{\dagger,\ddagger}$, Hengchang Guo$^{\ddagger,\ast}$,\footnotetext[1], Zhifei Zhang$^{\star}$, Wenxin Liu$^{\ddagger}$, Zhan Qin$^{\dagger}$, Kui Ren$^{\dagger,\sharp}$ \\
$^{\dagger}$School of Cyber Science and Technology, Zhejiang University, P. R. China\\
$^{\ddagger}$School of Cyber Science and Engineering, Wuhan University, P. R. China,\quad
$^{\star}$Adobe Research\\
$^{\sharp}$Key Laboratory of Blockchain and Cyberspace Governance of Zhejiang Province, P. R. China\\
{\tt\small \{zhibowang, qinzhan, kuiren\}@zju.edu.cn, \{hc\_guo, wxliu111\}@whu.edu.cn, zzhang@adobe.com}
}

\maketitle
\footnotetext[1]{Hengchang Guo is the corresponding author.}
\ificcvfinal\thispagestyle{empty}\fi

\begin{abstract}
Transferability of adversarial examples is of central importance for attacking an unknown model, which facilitates adversarial attacks in more practical scenarios, \eg, black-box attacks.
Existing transferable attacks tend to craft adversarial examples by indiscriminately distorting features to degrade prediction accuracy in a source model without aware of intrinsic features of objects in the images.
We argue that such brute-force degradation would introduce model-specific local optimum into adversarial examples, thus limiting the transferability.
By contrast, we propose the Feature Importance-aware Attack ({\proposed}), which disrupts important object-aware features that dominate model decisions consistently.
More specifically, we obtain feature importance by introducing the aggregate gradient, which averages the gradients with respect to feature maps of the source model, computed on a batch of random transforms of the original clean image.  The gradients will be highly correlated to objects of interest, and such correlation presents invariance across different models. Besides, the random transforms will preserve intrinsic features of objects and suppress model-specific information.
Finally, the feature importance guides to search for adversarial examples towards disrupting critical features, achieving stronger transferability.
Extensive experimental evaluation demonstrates the effectiveness and superior performance of the proposed {\proposed}, \ie, improving the success rate by 9.5\% against normally trained models and 12.8\% against defense models as compared to the state-of-the-art transferable attacks. Code is available at: https://github.com/hcguoO0/FIA
\end{abstract}

\section{Introduction}\label{sec:intro}

Deep neural networks (DNNs) have achieved superior performance in many vision tasks, \eg, image classification~\cite{DBLP:conf/nips/KrizhevskySH12,DBLP:conf/eccv/HeZRS16}, object detection~\cite{DBLP:conf/iccv/Girshick15,DBLP:conf/nips/RenHGS15}, semantic segmentation~\cite{DBLP:journals/pami/ChenPKMY18,DBLP:conf/cvpr/LongSD15}, face recognition~\cite{DBLP:conf/cvpr/TaigmanYRW14,DBLP:conf/cvpr/SunWT15}, \etc.
However, despite the impressive progress, recent studies showed that DNNs are vulnerable to adversarial examples~\cite{DBLP:journals/corr/SzegedyZSBEGF13} which are crafted by adding carefully designed perturbations to fool DNNs.
Adversarial attacks have raised great concern for DNN-based applications, especially in safety- and security-sensitive areas like autonomous driving.
Meanwhile, adversarial examples also play an important role in investigating the internal drawbacks of neural networks and improving their robustness.

\begin{figure}[t]
  \centering
  \includegraphics[width=.9\columnwidth]{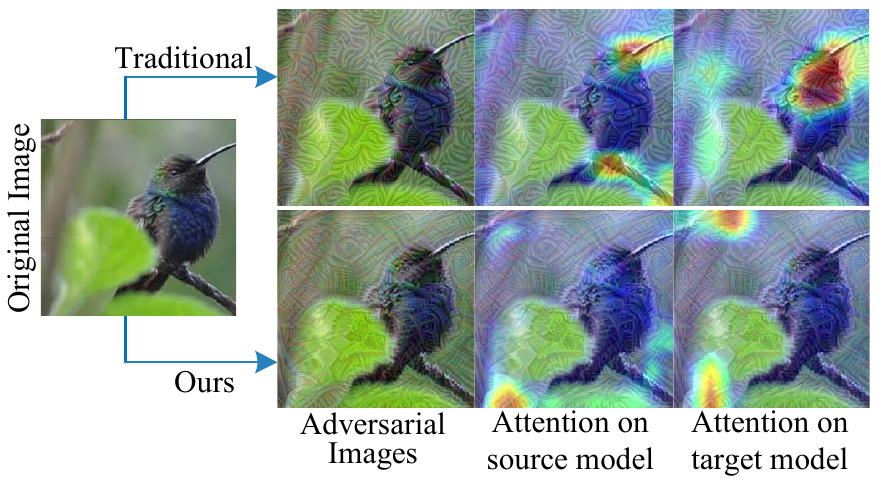}
  \caption{Comparison of traditionally indiscriminate attacks (top row) and our feature importance-aware attacks (bottom row). Adversarial images are generated on the source model (VGG16) and used to attack the target model (Inception-V3). Our attacks will suppress important features and promote trivial features, thus significantly defocusing/misleading the models as visualized by the attention maps, \ie, our adversarial example causes the source and target models not only failed to capture the important object but also focus on trivial regions.}
  \label{fig:intro}
  \vspace{-6mm}
\end{figure}

Many works~\cite{DBLP:journals/corr/SzegedyZSBEGF13,DBLP:journals/corr/GoodfellowSS14,DBLP:conf/iclr/KurakinGB17a,DBLP:conf/sp/Carlini017,DBLP:conf/ijcai/XiaoLZHLS18,DBLP:conf/icml/UesatoOKO18,DBLP:conf/cvpr/XieZZBWRY19} have been proposed to generate adversarial examples, which can be divided into two categories, \ie, white-box attacks vs. black-box attacks, according to the knowledge owned by attackers.
With the progress of adversarial attacks, the more challenging black-box attacks have attracted more attention.
A common type of black-box attacks~\cite{DBLP:conf/icml/UesatoOKO18,DBLP:conf/icml/IlyasEAL18,DBLP:conf/iclr/BrendelRB18} is to craft adversarial examples by estimating gradients based on queried information (\eg, probability vectors and hard labels), which is referred to as the query-based attack. Those query-based attacks may be impractical in the real world since excessive queries would not be allowed.
By contrast, another typical black-box attack, called transfer-based attack, relies on the cross-model transferability of adversarial examples~\cite{DBLP:conf/iclr/LiuCLS17} (\ie, adversarial examples crafted on one model could successfully attack other models for the same task), which is more practical and flexible.

However, adversarial examples crafted by traditional attacking methods (\eg, FGSM~\cite{DBLP:journals/corr/GoodfellowSS14}, BIM~\cite{DBLP:conf/iclr/KurakinGB17a}, \etc) often exhibit weak transferability due to overfitting to the source model. Therefore, some studies attempted to alleviate such overfitting by introducing extra operations during the optimization to improve transferability, \eg, random transformation~\cite{DBLP:conf/cvpr/XieZZBWRY19}, translation operation~\cite{DBLP:conf/cvpr/DongPSZ19}.
Recently, \cite{DBLP:conf/eccv/ZhouHCTHGY18,DBLP:conf/iccv/GaneshanSR19,DBLP:conf/cvpr/LuJWLCCV20} performed attacks in the intermediate layers directly to enhance transferability.
Instead of disturbing the output layer, these feature-level attacks maximize internal feature distortion and achieve higher transferability.
However, existing methods generate adversarial examples by indiscriminately distorting features without aware of the intrinsic features of objects in the images, thus easily trapped into model-specific local optimum.
Because classifiers tend to extract any available signal to maximize classification accuracy, even those imperceptible noises implied in the images~\cite{DBLP:conf/nips/IlyasSTETM19}, the model will learn extra ``noisy'' features together with intrinsic features of objects, while the ``noisy'' features are treated equally with object-related features to support the final decision, and such ``noisy'' features will be model-specific.
Therefore, the adversarial examples crafted by existing methods tend to distort such model-specific features, thus overfitting to the source model and hindering the transferability of the adversarial examples.

This paper proposes a Feature Importance-aware Attack ({\proposed}), which significantly improves the transferability of adversarial examples by disrupting the important object-aware features that are supposed to dominate the decision of different models.
Against model-specific features, we introduce aggregate gradient, which will effectively suppress model-specific features while at the same time providing object-aware importance of the features.
As illustrated in Fig.~\ref{fig:intro}, compared to traditionally indiscriminate attacks, the adversarial image from the proposed \proposed{} significantly defocuses the models, \ie, failed to capture the important features of the object. Meanwhile, the models are misled to focus on those trivial areas.
More specifically, random transformations (we adopt random pixel dropping) are first applied to the original images. Since the transformed images will preserve the spatial structure and texture but variating non-semantic details, the features from them will be consistent on the object-aware features but fluctuated on non-object (\ie, model-specific ``noisy'') features. With respect to these features, gradients are averaged to statistically suppress those fluctuated model-specific features. Meanwhile, object-aware/important features are preserved to guide the generation of more transferable adversarial examples since the feature importance is highly correlated to objects of interest and consistent across different models.

Our main contributions are summarized as follows.
\begin{itemize}
  \item We propose Feature Importance-aware Attack ({\proposed}) that enhances the transferability of adversarial examples by disrupting the critical object-aware features that dominate the decision of different models.
  \item We analyze the rationale behind the relatively low transferability of existing works, \ie, overfitting to model-specific ``noisy'' features, against which we introduce aggregate gradient to guide the generation of more transferable adversarial examples.
  \item Extensive experiments on diverse classification models demonstrate the superior transferability of adversarial examples generated by the proposed \proposed{} as compared to state-of-the-art transferable attacking methods.
\end{itemize}

\section{Related Work}\label{sec:related}

Since Szegedy~\etal~\cite{DBLP:journals/corr/SzegedyZSBEGF13} demonstrated the existence of adversarial examples, many adversarial attack algorithms~\cite{DBLP:journals/corr/GoodfellowSS14,DBLP:conf/iclr/KurakinGB17a,DBLP:conf/sp/Carlini017,DBLP:conf/icml/UesatoOKO18,DBLP:conf/icml/IlyasEAL18,DBLP:conf/iclr/BrendelRB18} have been proposed to discuss the vulnerability of neural networks.
In this work, we mainly focus on the transfer-based attacks which utilize the transferability of adversarial examples to perform black-box attack, \ie,  using adversarial examples crafted on one model to attack other models.

\begin{figure*}[!t]
  \centering
  \includegraphics[width=1.8\columnwidth]{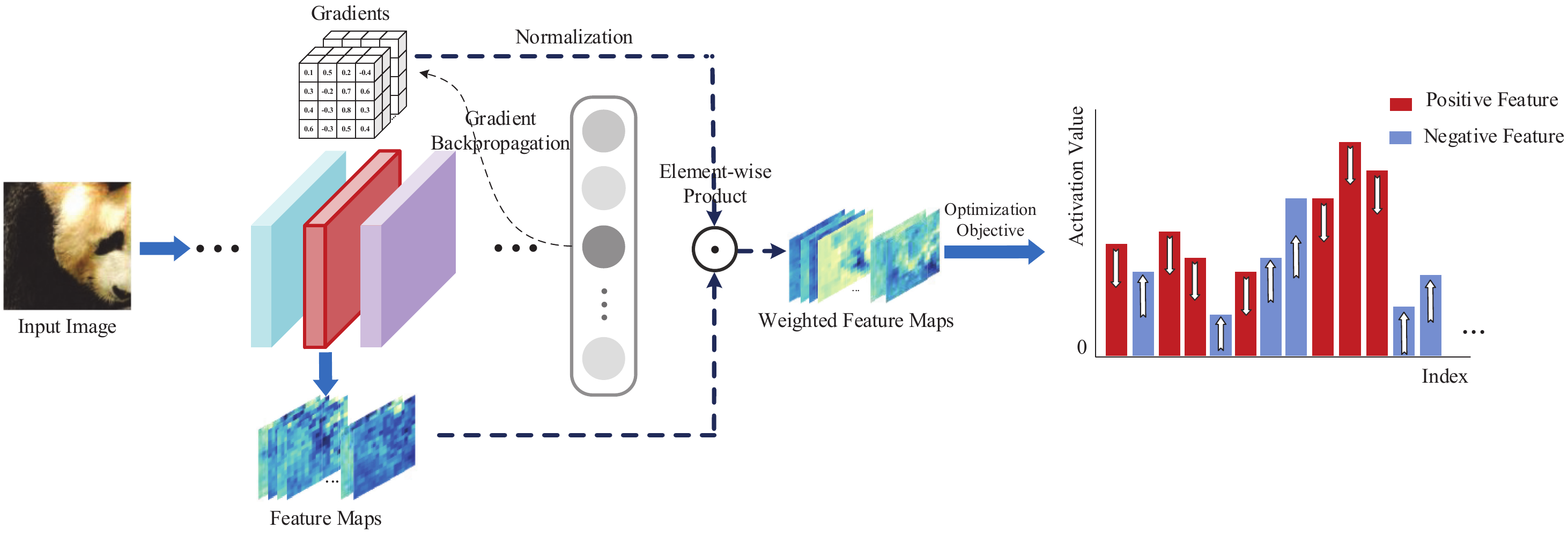}
  \caption{Overview of Feature Importance-aware Attack ({\proposed}). Given an input image, feature maps are extracted from an intermediate layer (red block) of the source classification model. Then, the gradients back propagated from the output to the feature maps are calculated to serve as the feature importance. After element-wise product between the feature maps and normalized gradients (\ie, feature importance), the weighted feature maps could be optimized by specifically suppressing positive/important features and promoting negative/trivial features, achieving higher transferable adversarial examples.}
  \label{fig:framework}
  \vspace{-5mm}
\end{figure*}

There are many works aiming to improve the transferability of adversarial examples.
Dong~\etal~\cite{DBLP:conf/cvpr/DongLPS0HL18} adopted momentum in iterative algorithms to stabilize updates and avoid poor local optimum, while Lin~\etal~\cite{DBLP:conf/iclr/LinS00H20} introduced Nesterov accelerated gradient to further enhance transferability.
Xie~\etal~\cite{DBLP:conf/cvpr/XieZZBWRY19} applied image transformation (\ie, randomly resizing and padding) to the inputs at each iteration to improve transferability.
Gao~\etal~\cite{DBLP:conf/eccv/GaoZSLS20} crafted patch-wise noise instead of pixel-wise noise to enhance the transferability of adversarial examples.
Dong~\etal~\cite{DBLP:conf/cvpr/DongPSZ19} proposed translation-invariant attack by optimizing perturbations over translated images which leads to higher transferability against defense models.

Instead of disrupting the output layer, several works proposed to attack internal features.
Zhou~\etal~\cite{DBLP:conf/eccv/ZhouHCTHGY18} first demonstrated that maximizing the feature distance between natural images and their adversarial examples in the intermediate layers can boost the transferability. Naseer~\etal~\cite{naseer2018task} also concluded that neural representation distortion does not suffer from the overfitting problem, and can exhibit cross-architecture, cross-dataset and cross-task transferability. Huang~\etal~\cite{DBLP:conf/iccv/HuangKGHBL19} fine-tuned an existing adversarial example for greater transferability by increasing its perturbation on a pre-specified layer from the source model. Ganeshan~\etal~\cite{DBLP:conf/iccv/GaneshanSR19} proposed a principled manner to inflict severe damage to feature representation, hence obtaining higher transferability.
Our proposed method also falls into this category, and the key difference is that our method considers the feature importance and disrupts the critical object-aware features that dominate the decision across different models, while existing methods distort features indiscriminately.

\section{Preliminaries}\label{sec:pre}
Assume a classification model $f_\theta: x \mapsto y$, where $x$ and $y$ denotes the clean image and true label, respectively, and $\theta$ indicates the parameters of the model.  We aim to generate an adversarial example $x^{adv}=x+\epsilon$, which is distorted by carefully designed perturbation $\epsilon$ but will mislead the classifier, \ie, $f_\theta(x^{adv}) \neq y$. Typically, $\ell_p$-norm is commonly adopted to regularize the perturbation. Therefore, the generation of adversarial examples can be formulated as an optimization problem as shown in the following.
\begin{equation}
\arg \underset{x^{adv}}{\max }\; J\left(x^{adv}, y\right), \;
s.t.\;\left\|x-x^{adv}\right\|_p \leq \epsilon,
\label{equ:definition}
\end{equation}
where the loss function $J(\cdot, \cdot)$ measures the distance between true and predicted labels (\ie, cross-entropy), and $p=\infty$ in this work. Many methods have been proposed to solve the above optimization problem, \eg, Fast Gradient Sign Method (FGSM)~\cite{DBLP:journals/corr/GoodfellowSS14}, Basic Iterative Method (BIM)~\cite{DBLP:conf/iclr/KurakinGB17a}, Momentum Iterative Method (MIM)~\cite{DBLP:conf/cvpr/DongLPS0HL18}, \etc.
However, optimizing Eq.~\ref{equ:definition} requires explicit access to the parameters of $f_\theta$, while this is impractical in black-box attacking. Therefore, a feasible solution is to optimize on an analogous model $f_\phi$ (\ie, the source model) with accessible parameters $\phi$, thus generating highly transferable adversarial examples to attack the target model $f_\theta$.

\begin{figure}
  \centering
  \includegraphics[width=0.85\columnwidth]{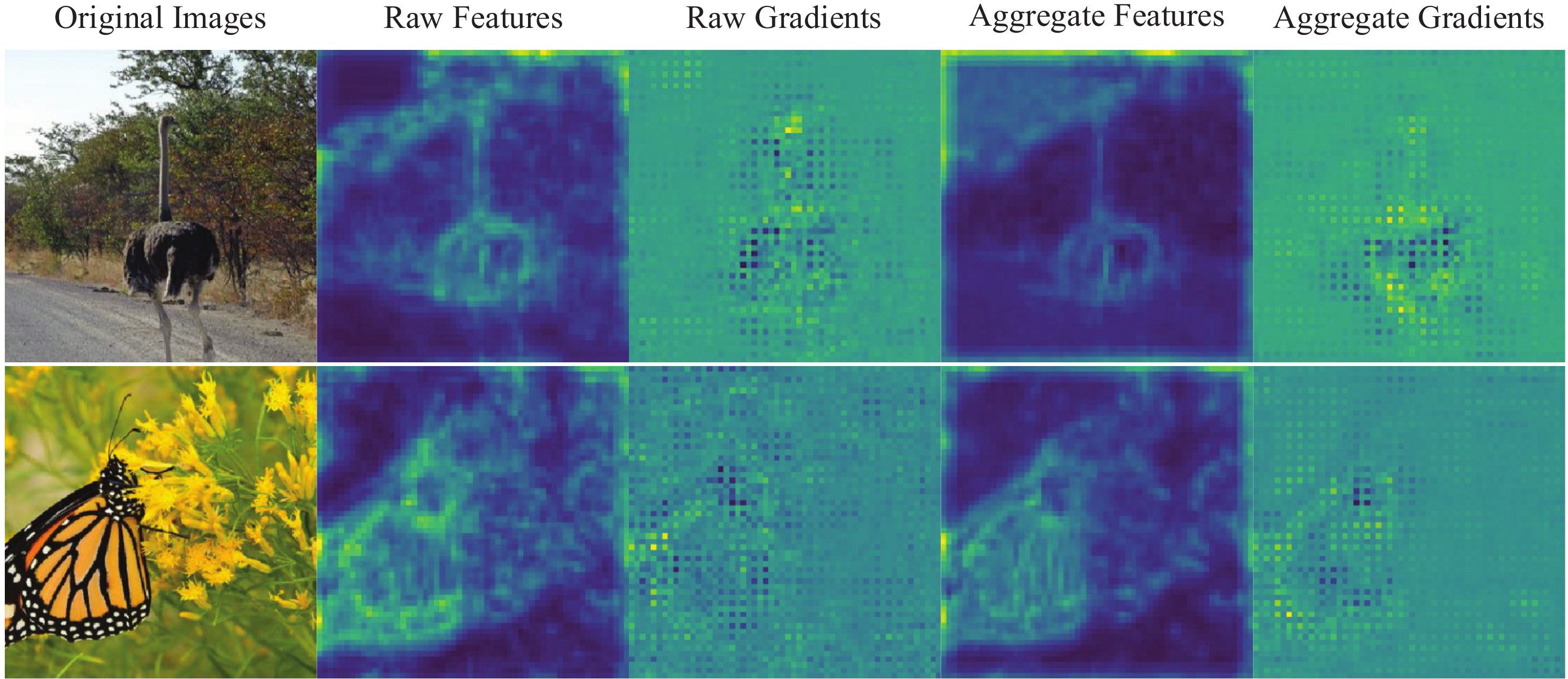}
  \caption{Visualization of feature maps and corresponding gradients at the layer Conv3\_3 of VGG16. Raw features and gradients are obtained from the original images, and aggregate features and gradients are obtained from multiple transforms (\ie, random pixel dropping) of the original images.}
  \label{fig:grad}
  \vspace{-6mm}
\end{figure}

\section{Feature Importance-aware Attack}
Empirical studies from most DNN-based classifiers, \eg, Inception~\cite{DBLP:conf/cvpr/SzegedyVISW16} and ResNet~\cite{DBLP:conf/cvpr/HeZRS16}, have shown that deep models tend to extract semantic features, which are object-aware discriminative and thus effectively boost classification accuracy. Intuitively, disrupting those object-aware features that dominate the decision of all models could benefit the transferability of adversarial examples.
However, different networks also extract exclusive features to better fit themselves to the data domain, which results in model-specific feature representation.
Without awareness of such characteristics, existing adversarial attacks tend to craft adversarial examples by indiscriminately distorting features against a source model, thus trapped into model-specific local optimum and significantly degrading the transferability.
Therefore, avoiding such local optimum is a key to transferability. More specifically, the generation of adversarial examples needs to be guided by model-agnostic critical features from the source model, which is referred to as feature importance.
Fig.~\ref{fig:framework} overviews the proposed feature importance-aware transferable attack (\proposed), where aggregate gradient (detailed in Section~\ref{sec:fim}) can effectively avoid local optimum and represent transferable feature importance. Then, aggregate gradient serves as weights in the optimization to distort important features (discussed in Section~\ref{subsec:alg}), \ie, reducing features with positive weights and increasing those corresponding to negatives.

\subsection{Feature Importance by Aggregate Gradient}
\label{sec:fim}

\begin{figure}
  \centering
  \includegraphics[width=0.85\columnwidth]{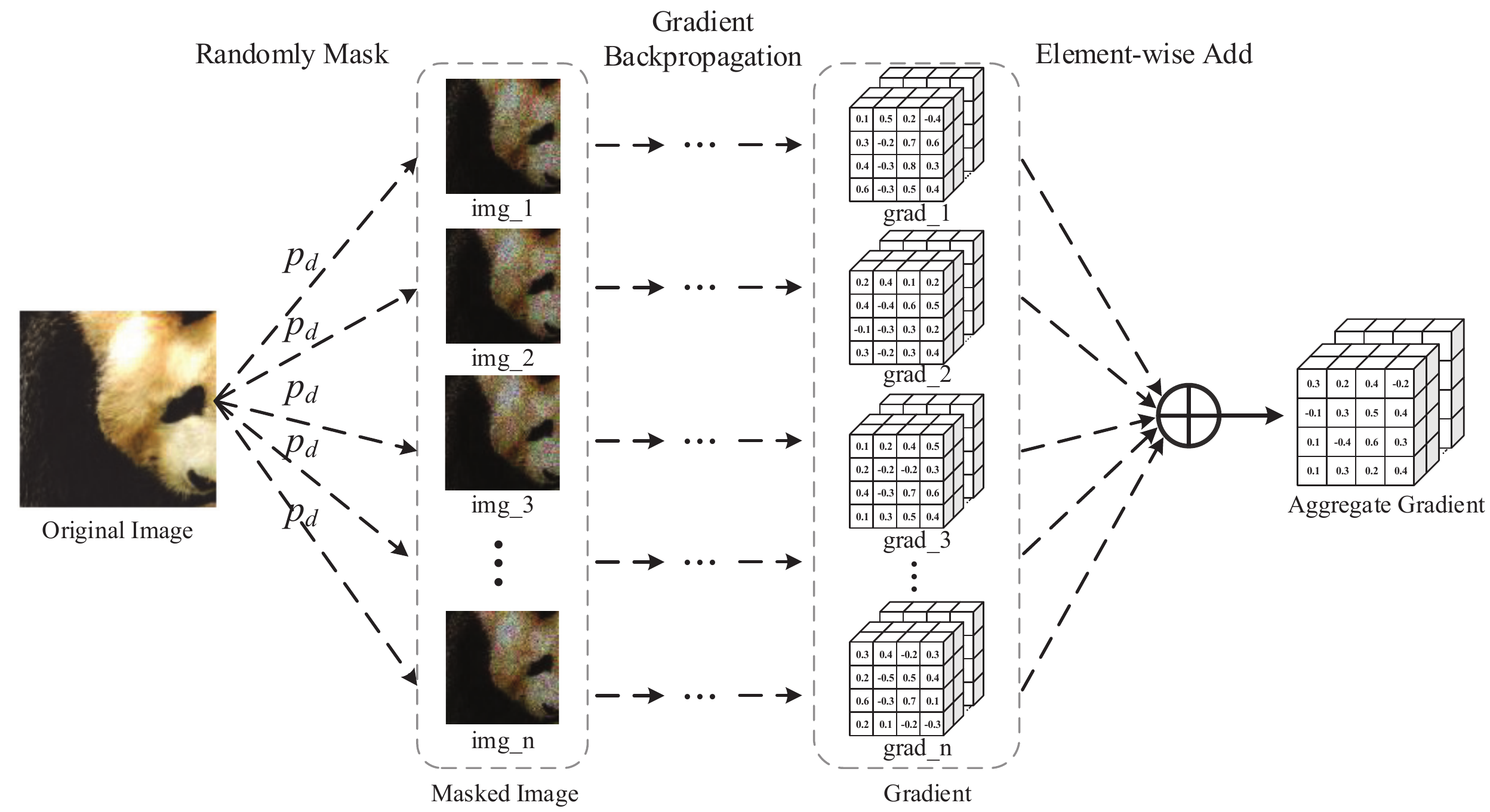}
  \caption{Illustration of the aggregate gradient. The gradients are obtained from multiple random masked images, and the final aggregate gradient (\ie, feature importance) is represented by averaging these gradients.}
  \label{fig:ensemble}
  \vspace{-5mm}
\end{figure}

For simplicity, let $f$ denote the source model, and the feature maps from the $k$-th layer are expressed as $f_k(x)$. Since feature importance is proportional to how the features contribute to the final decision, an intuitive strategy is to obtain the gradient w.r.t. $f_k(x)$ as written in the following.
\begin{equation}
\Delta_k^x =\frac{\partial l\left(x, t\right)}{\partial f_{k}(x)},
\label{equ:gradient}
\end{equation}
where $l(\cdot, \cdot)$ denotes the logit output with respect to the true label $t$.
However, the raw gradient $\Delta_k^x$ would carry model-specific information. As shown in Fig.~\ref{fig:grad}, the raw gradient maps and raw feature maps are both visually noisy, \ie, pulses and large gradient on non-object regions, which may be caused by the model-specific solution space.

To suppress model-specific information, we propose the aggregate gradient, which aggregates gradients from randomly transformed $x$ as shown in Fig.~\ref{fig:ensemble}. The transformation is supposed to distort image details but preserve the spatial structure and general texture. Since semantically object-aware/important features/gradients are robust to such transformation but model-specific ones are vulnerable to the transforms, those robust/transferable features/gradients will be highlighted after aggregation, while the others would be neutralized. In this paper, we adopt random pixel dropping (\ie, randomly mask) with the probability $p_d$. Therefore, the aggregate gradient can be expressed in the following.
\begin{equation}
\bar{\Delta}_k^{x} = \frac{1}{C}\sum_{n=1}^N \Delta_k^{x \odot M^n_{p_d}}, \; M_{p_d} \sim \text{Bernoulli}(1-p_d),
\label{equ:agg_grad}
\end{equation}
where the $M_{p_d}$ is a binary matrix with the same size as $x$, and $\odot$ denotes the element-wise product. The normalizer $C$ is obtained by $\ell_2$-norm on the corresponding summation term. The ensemble number $N$ indicates the number of random masks applied to the input $x$. Fig.~\ref{fig:ensemble} illustrates the process of gradient aggregation.
The aggregate gradient $\bar{\Delta}_k^{x}$ highlights the regions of robust and critical object-aware features that can guide the adversarial example towards the more transferable direction.
Fig.\ref{fig:grad} visualizes the aggregate gradient, which is cleaner and focuses more on the objects as compared to the raw gradient, providing better feature importance in the transferable perspective.

\begin{algorithm}[t]
	\caption{Feature Importance-aware Attack}
	\KwIn{The original clean image $x$, classification model $f$, intermediate layer $k$, drop probability $p_d$ and ensemble number $N$ in aggregate gradient, max perturbation $\epsilon$, and the number of iteration $T$.}
	\KwOut{The adversarial image $x^{adv}$.}
	\textbf{Initialize:}$\Delta=0$, $g_{0}=0$, $\mu=1$, $\alpha=\epsilon/T$\\
	Obtain aggregate gradient:\\
	\For{ n = 0 to N - 1}
	{
		\quad\quad$\displaystyle\Delta$ = $\Delta$ + $\Delta_{k}^{x \odot M^n_{p_d}}$ \\
	}

	\quad\quad $\displaystyle\Delta = \frac{\Delta}{\|\Delta\|_2}$ \\
	
	Construct optimization objective: \\
	\quad\quad $\displaystyle\mathcal{L}(x^{adv})$ = $\sum \left(\Delta \odot f_{k}(x^{adv}) \right)$ \\ 
	Update $x^{adv}$ by momentum iterative method: \\
	\For{ t =0 to T - 1}
	{
		\quad\quad$\displaystyle g_{t+1}=\mu \cdot g_{t} + \frac{\nabla_{x} \mathcal{L}\left(x_{t}^{adv}\right)}{\left\|\nabla_{x} \mathcal{L}\left(x_{t}^{adv}\right)\right\|_{1}}$ \\
		\quad\quad$\displaystyle x_{t+1}^{adv}=\operatorname{Clip}_{x,\epsilon}\left\{x_{t}^{adv}-\alpha \cdot \operatorname{sign}\left(g_{t+1}\right)\right\}$ \\
	}
	\Return{$x^{adv}_{T}$}
	\label{alg:our}

\end{algorithm}

\begin{table*}[h]\small
  \centering
  {
  \tabcolsep 10pt
  \caption{Success rate of different attacks against normally trained models. The first column shows source models, and the first row lists target models. \proposed{} is our method and \proposed{}+PIDIM is the combination of \proposed{} and PIDIM. ``*'' indicates white-box attack since the target model is the source model, and the best results are highlighted in bold.}
  \label{tab:compareNor}
  \renewcommand\arraystretch{0.85}
  \begin{tabular}{c|>{\columncolor{mygray}}c|>{\columncolor{mygray}}c|>{\columncolor{mygray}}c|>{\columncolor{mygray}}c|>{\columncolor{mygray}}c|>{\columncolor{mygray}}c|>{\columncolor{mygray}}c|>{\columncolor{mygray}}c}
    \toprule
    \rowcolor{white}
     & Attack &  Inc-v3 & Inc-v4 & IncRes-v2 & Res-50 & Res-152 & Vgg-16 & Vgg-19 \\
     \midrule
     \rowcolor{white}
     &  MIM & \textbf{100.0\%*} & 41.6\% & 38.8\% & 33.1\% & 29.7\% & 38.6\% & 38.3\%\\
     \rowcolor{white}
     & DIM & 99.6\%* & 64.6\% & 59.6\% & 40.7\% & 36.3\% & 47.6\% & 46.4\% \\
     \rowcolor{white}
     & PIM & 97.9\%* & 55.8\% & 51.5\% & 53.3\% & 46.3\% & 61.6\% & 60.5\%\\
     \rowcolor{white}
     & PIDIM & 98.1\%* & 70.5\% & 66.4\% & 61.8\% & 56.3\% & 57.7\% & 56.0\%\\
     \rowcolor{white}
     & NRDM & 90.2\%* & 60.3\% & 51.2\% & 42.6\% & 31.8\% & 38.5\% & 39.2\%\\
     \rowcolor{white}
     & FDA & 81.3\%* & 42.4\% & 36.4\% & 29.6\% & 25.5\% & 31.5\% & 30.5\%\\
     & {\proposed} & 98.3\%* & 83.5\% & 80.6\% & 70.4\% & 64.9\% & 71.4\% & 73.3\%\\
    \multirow{-8}*{Inc-v3}  & {\proposed}+PIDIM & 98.8\%* & \textbf{87.8\%} & \textbf{85.7\%} & \textbf{79.7\%} & \textbf{74.4\%} & \textbf{82.4\%} & \textbf{84.1\%}\\
    \midrule
    \rowcolor{white}
    & MIM & 60.2\% & 52.5\% & 99.3\%* & 40.1\% & 36.1\% & 46.9\% & 43.8\%\\
    \rowcolor{white}
    & DIM & 75.2\% & 71.3\% & 97.1\%* & 50.9\% & 43.7\% & 51.5\% & 51.4\% \\
    \rowcolor{white}
    & PIM & 66.8\% & 62.9\% & \textbf{99.6\%*} & 56.2\% & 50.8\% & 64.4\% & 63.5\%\\
    \rowcolor{white}
    & PIDIM & 80.5\% & 78.0\% & 98.5\%* & 56.6\% & 50.1\% & 62.5\% & 62.6\%\\
    \rowcolor{white}
    & NRDM & 70.6\% & 67.3\% & 77.0\%* & 56.4\% & 45.9\% & 50.2\% & 51.7\%\\
    \rowcolor{white}
    & FDA & 69.2\% & 67.8\% & 78.3\%* & 52.8\% & 39.9\% & 46.2\% & 44.4\%\\
    & {\proposed} & 81.1\% & 77.5\% & 89.2\%* & 71.8\% & 68.9\% & 71.4\% & 71.4\%\\
    \multirow{-8}*{IncRes-v2} & {\proposed}+PIDIM & \textbf{84.2\%} & \textbf{79.7\%} & 91.6\%* & \textbf{79.0\%} & \textbf{78.4\%} & \textbf{80.6\%} & \textbf{79.9\%}\\
    \midrule
    \rowcolor{white}
     & MIM & 57.2\% & 48.2\% & 45.7\% & 90.6\% & 99.8\%* & 72.8\% & 72.9\%\\
     \rowcolor{white}
     & DIM & 80.3\% & 72.2\% & 72.6\% & 95.0\% & 99.9\%* & 88.4\% & 88.0\% \\
     \rowcolor{white}
     & PIM & 66.0\% & 56.4\% & 51.1\% & 92.3\% & \textbf{100.0\%*} & 83.2\% & 82.5\%\\
     \rowcolor{white}
     & PIDIM & 82.2\% & 76.6\% & 77.0\% & 96.7\% & 99.8\%* & 91.2\% & 89.9\%\\
     \rowcolor{white}
     & NRDM & 64.5\% & 59.1\% & 51.2\% & 87.7\% & 95.4\%* & 79.3\% & 79.3\%\\
     \rowcolor{white}
     & FDA & 60.7\% & 52.3\% & 48.0\% & 85.0\% & 95.3\%* & 75.0\% & 75.0\%\\
     & {\proposed} & 85.3\% & 81.1\% & 77.8\% & 96.8\% & 99.5\%* & 91.5\% & 91.5\%\\
    \multirow{-8}*{Res-152} & {\proposed}+PIDIM & \textbf{90.3\%} & \textbf{85.9\%} & \textbf{85.6\%} & \textbf{98.2\%} & 99.5\%* & \textbf{95.8\%} & \textbf{95.7\%}\\
    \midrule
    \rowcolor{white}
     & MIM & 80.3\% & 81.1\% & 74.6\% & 89.3\% & 84.4\% & \textbf{100.0\%*} & 96.5\%\\
     \rowcolor{white}
     & DIM & 87.2\% & 87.0\% & 80.9\% & 92.0\% & 87.8\% & 99.8\%* & 98.9\% \\
     \rowcolor{white}
     & PIM & 84.1\% & 82.0\% & 75.6\% & 91.1\% & 85.9\% & \textbf{100.0\%*} & 98.9\%\\
     \rowcolor{white}
     & PIDIM & 89.1\% & 89.5\% & 84.7\% & 93.8\% & 90.8\% & 99.9\%* & 98.8\%\\
     \rowcolor{white}
     & NRDM & 73.6\% & 72.8\% & 57.1\% & 77.5\% & 73.0\% & 93.2\%* & 91.1\%\\
     \rowcolor{white}
     & FDA & 76.1\% & 76.7\% & 64.0\% & 81.7\% & 78.7\% & 95.7\%* & 95.7\%\\
     & {\proposed}& 95.7\% & 95.6\% & 92.3\% & 97.3\% & 95.3\% & 99.8\%* & 99.6\%\\
   	\multirow{-8}*{Vgg-16} & {\proposed}+PIDIM & \textbf{97.6\%} & \textbf{97.5\%} & \textbf{93.8\%} & \textbf{98.2\%} & \textbf{96.4\%} & 99.8\%* & \textbf{99.8\%}\\
    \bottomrule
  \end{tabular}}
  \vspace{-5mm}
\end{table*}

\subsection{Attack Algorithm}
\label{subsec:alg}

Utilizing the aforementioned feature importance (\ie, aggregate gradient $\bar{\Delta}_k^{x}$), we design the loss function Eq.~\ref{equ:loss} to guide the generation of adversarial example $x^{adv}$ by explicitly suppress important features.
For simplicity, we denote $\bar{\Delta}_k^{x}$ as $\Delta$ in the rest of this paper.
\begin{equation}
\mathcal{L}(x^{adv}) =  \sum\left( \Delta \odot f_{k}(x^{adv}) \right).
\label{equ:loss}
\end{equation}

Intuitively, the important features will yield relatively higher intensity in $\Delta$, which indicates the efforts of correcting the features to approach the true label, and the sign of $\Delta$ provides the correcting direction. The objective of generating transferable adversarial examples is to decrease important features with positive $\Delta$ and increase those corresponding to negative $\Delta$. Therefore, it is straightforward to achieve this goal by minimizing Eq.~\ref{equ:loss}. Finally, substitute the Eq.~\ref{equ:loss} into Eq.~\ref{equ:definition}, we get the proposed objective for feature importance-aware transferable adversarial attack.
\begin{equation}
\arg\underset{x^{adv}}{ \min } \; \mathcal{L}\left(x^{adv}\right), \; s.t. \; \left\|x-x^{adv}\right\|_{\infty} \leq \epsilon.
\label{equ:obj}
\end{equation}

There are many existing gradient-based attack methods aiming to solve the above objective function Eq.~\ref{equ:obj}, \eg, BIM~\cite{DBLP:conf/iclr/KurakinGB17a}, MIM~\cite{DBLP:conf/cvpr/DongLPS0HL18}, etc. Given the superior performance of MIM, we adopt this method to solve Eq. \ref{equ:obj}, and the details are shown in Algorithm~\ref{alg:our}.

\subsection{Drawback of Related Attacks}

It is worth to further emphasize the advantage of feature importance awareness over related feature-based attacks, \ie,  NRDM~\cite{naseer2018task} and FDA~\cite{DBLP:conf/iccv/GaneshanSR19}. For better illustration, their loss functions are written in Eq.~\ref{equ:nrdm} and \ref{equ:fda}, respectively
\begin{equation}
\mathcal{L}_{NRDM} = \left\| f_{k}(x^{adv})-f_{k}(x)\right\|_{2}, \\
\label{equ:nrdm}
\end{equation}
where $\ell_2$-norm is adopted to simply measure the feature distortion, which indiscriminately disturbs the features.
\begin{equation}
\begin{aligned}
\label{equ:fda}
\mathcal{L}_{FDA} = &\log (\mathcal{D}({f_{k}(x^{adv})\mid f_{k}(x)<C_{k}(h, w)})) \\
            - & \log (\mathcal{D}({f_{k}(x^{adv}) \mid f_{k}(x)>C_{k}(h, w)}))
\end{aligned}
\end{equation}
where $\mathcal{D}(\cdot)$ is the $\ell_2$-norm, and $C_{k}(h, w)$ represents the mean activation values across channels.

From the objective functions, NRDM simply optimizes the feature distortion between original images and adversarial images without any constraints.
For FDA, although it introduced a similar idea of utilizing feature activation to guide the optimization, \ie, features that support the ground truth should be suppressed, while those that do not support the ground truth should be enhanced. However, FDA uses the mean across channels as the distinguishable criterion, which cannot effectively avoid model-specific information.
By contrast, the proposed \proposed{} provides more intrinsic feature importance by the aggregate gradient, and hence achieving higher transferable adversarial examples. Quantitative comparison in the experimental evaluations demonstrate the superior performance of the proposed \proposed{}.

\section{Experimental Evaluation}

\begin{table*}[h]\small
  \centering
  {
  \caption{Success rate of different attacks against defense models. The first column shows source models, and the first row lists target models. \proposed{} is our method and \proposed{}+PITIDIM is the combination of \proposed{} and PITIDIM. The best results are highlighted in bold.}
  \label{tab:compareAdv}
  \renewcommand\arraystretch{0.85}
  \newcommand{\tabincell}[2]{\begin{tabular}{@{}#1@{}}#2\end{tabular}}
  \begin{tabular}{c|>{\columncolor{mygray}}c|>{\columncolor{mygray}}c|>{\columncolor{mygray}}c|>{\columncolor{mygray}}c|>{\columncolor{mygray}}c|>{\columncolor{mygray}}c}
    \toprule
    \rowcolor{white}
    &Attack&Adv-Inc-v3& Adv-IncRes-v2& Ens3-Inc-v3&Ens4-Inc-v3&Ens-IncRes-v2\\
    \midrule
       \rowcolor{white}
       & MIM & 22.9\% & 17.5\% & 15.4\% & 15.8\% & 7.8\% \\
       \rowcolor{white}
       & DIM & 26.0\% & 24.5\% & 17.8\% & 20.8\% & 10.0\% \\
       \rowcolor{white}
       & TIM & 32.0\% & 26.4\% & 30.1\% & 32.5\% & 22.4\%  \\
       \rowcolor{white}
       & PIM & 34.3\% & 30.2\% & 33.3\% & 38.4\% & 26.2\% \\
       \rowcolor{white}
       & TIDIM & 40.7\% & 37.1\% & 40.8\% & 42.3\% & 30.4\% \\
       \rowcolor{white}
       & PITIDIM & 41.6\% & 33.9\% & 43.1\% & 47.3\% & 31.4\% \\
       & {\proposed} & 54.5\% & 54.9\% & 43.9\% & 42.0\% & 23.5\% \\
       \multirow{-8}*{Inc-v3}& {\proposed}+PITIDIM & \textbf{64.8\%} & \textbf{59.0\%} & \textbf{62.5\%} & \textbf{63.2\%} & \textbf{50.9\%} \\
    \midrule
    \rowcolor{white}
     & MIM & 25.5\% & 29.9\% & 21.4\% & 22.7\% & 12.5\% \\
     \rowcolor{white}
     & DIM & 33.1\% & 42.9\% & 30.5\% & 29.7\% & 19.0\% \\
     \rowcolor{white}
     & TIM & 40.0\% & 43.5\% & 39.5\% & 41.5\% & 38.4\% \\
     \rowcolor{white}
     & PIM & 39.0\% & 35.3\% & 39.4\% & 42.2\% & 32.8\% \\
     \rowcolor{white}
     & TIDIM & 50.4\% & 55.7\% & 50.1\% & 49.5\% & 48.1\% \\
     \rowcolor{white}
     & PITIDIM & 53.8\% & 55.2\% & 54.7\% & 54.5\% & \textbf{50.6\%} \\
     & {\proposed} & 54.9\% & \textbf{56.8\%} & 46.9\% & 44.7\% & 37.4\% \\
     \multirow{-8}*{IncRes-v2}& {\proposed}+PITIDIM & \textbf{55.1\%} & 52.9\% & \textbf{54.9\%} & \textbf{56.2\%} & \textbf{50.6\%} \\
    \midrule
    \rowcolor{white}
       & MIM & 36.9\% & 34.8\% & 36.2\% & 37.4\% & 22.0\% \\
       \rowcolor{white}
       & DIM & 54.3\% & 54.6\% & 53.3\% & 50.4\% & 33.5\% \\
       \rowcolor{white}
       & TIM & 41.5\% & 37.5\% & 43.1\% & 47.6\% & 34.1\% \\
       \rowcolor{white}
       & PIM & 40.7\% & 38.9\% & 46.9\% & 51.8\% & 38.8\% \\
       \rowcolor{white}
       & TIDIM & 52.4\% & 48.6\% & 57.5\% & 61.1\% & 46.3\% \\
       \rowcolor{white}
       & PITIDIM & 51.9\% & 49.0\% & 58.6\% & 64.8\% & 47.9\% \\
       & {\proposed} & \textbf{70.1\%} & \textbf{66.7\%} & 61.4\% & 60.3\% & 41.7\% \\
       \multirow{-8}*{Res-152}& {\proposed}+PITIDIM & 66.3\% & 62.5\% & \textbf{69.6\%} & \textbf{72.7\%} & \textbf{61.4\%} \\
    \midrule
    \rowcolor{white}
      & MIM & 64.3\% & 61.1\% & 64.3\% & 64.3\% & 45.0\% \\
       \rowcolor{white}
       & DIM & 69.9\% & 66.2\% & 70.3\% & 67.8\% & 49.9\% \\
       \rowcolor{white}
       & TIM & 52.8\% & 46.2\% & 55.1\% & 55.3\% & 41.6\% \\
       \rowcolor{white}
       & PIM & 51.9\% & 43.2\% & 50.2\% & 56.3\% & 39.9\% \\
       \rowcolor{white}
       & TIDIM & 59.1\% & 48.2\% & 59.6\% & 60.3\% & 47.9\% \\
       \rowcolor{white}
       & PITIDIM & 51.0\% & 44.6\% & 55.6\% & 60.7\% & 43.1\% \\
       & {\proposed} & \textbf{87.8\%} & \textbf{86.3\%} & \textbf{85.6\%} & \textbf{86.0\%} & \textbf{70.8\%} \\
      \multirow{-8}*{Vgg-16} & {\proposed}+PITIDIM & 74.7\% & 71.4\% & 77.3\% & 80.1\% & 67.0\% \\
    \bottomrule
  \end{tabular}}
  \vspace{-5mm}
\end{table*}

\subsection{Experiment Setup}
\label{sec:setting}

\noindent\textbf{Dataset:} For fair comparison, we follow the previous works~\cite{DBLP:conf/cvpr/DongPSZ19,DBLP:conf/eccv/GaoZSLS20} to use the ImageNet-compatible dataset~\cite{nips_competition}, which consists of 1000 images and was used for NIPS 2017 adversarial competition.

\noindent\textbf{Target Models:} The proposed \proposed{} is validated on twelve state-of-the-art classification models, including seven normally trained models\footnote{\tiny\url{https://github.com/tensorflow/models/tree/master/research/slim}} and five adversarially trained models\footnote{\tiny\url{https://github.com/tensorflow/models/tree/archive/research/adv_imagenet_models}} (\ie, defense models).
The normally trained models are Inception-V3 (Inc-v3)~\cite{DBLP:conf/cvpr/SzegedyVISW16}, Inception-V4 (Inc-v4)~\cite{DBLP:conf/aaai/SzegedyIVA17}, Inception-ResNet-V2 (IncRes-v2)~\cite{DBLP:conf/aaai/SzegedyIVA17}, ResNet-V1-50 (Res-50)~\cite{DBLP:conf/cvpr/HeZRS16}, ResNet-V1-152 (Res-152)~\cite{DBLP:conf/cvpr/HeZRS16}, VGG16 (Vgg-16)~\cite{DBLP:journals/corr/SimonyanZ14a}, and VGG19 (Vgg-19)~\cite{DBLP:journals/corr/SimonyanZ14a}.
With adversarial training~\cite{DBLP:conf/iclr/KurakinGB17,DBLP:conf/iclr/TramerKPGBM18}, the corresponding defense models are Adv-Inc-v3, Adv-IncRes-v2, Ens3-Inc-v3, Ens4-Inc-v3, and Ens-IncRes-v2.

\noindent\textbf{Baseline Attacks:} To demonstrate the effectiveness of the proposed \proposed{}, we compare it to diverse state-of-the-art attack methods, \eg, MIM~\cite{DBLP:conf/cvpr/DongLPS0HL18}, DIM~\cite{DBLP:conf/cvpr/XieZZBWRY19}, TIM\cite{DBLP:conf/cvpr/DongPSZ19}, PIM~\cite{DBLP:conf/eccv/GaoZSLS20}, as well as combined versions of these methods \ie, TIDIM~\cite{DBLP:conf/cvpr/DongPSZ19}, PIDIM~\cite{DBLP:conf/eccv/GaoZSLS20} and PITIDIM~\cite{DBLP:conf/eccv/GaoZSLS20}.
In addition, recent feature-level attacks are also involved, \ie, NRDM~\cite{naseer2018task}, FDA~\cite{DBLP:conf/iccv/GaneshanSR19}.

\noindent\textbf{Parameter Settings:} In all experiments, the maximum perturbation $\epsilon=16$, the iteration $T=10$, and the step size $\alpha=\epsilon/T=1.6$ (recap Algorithm~\ref{alg:our}).
For the baseline attacks, the transform probability is 0.7 in DIM, and the kernel size is 15 for TIM. Since settings of PIM will vary with target models and the ways of method combination, we will specifically detail its settings (\ie, the amplification factor $\beta$, project factor $\gamma$, and project kernel size $k_{w}$) in each related experiment.
In the proposed \proposed{}, the drop probability $p_d=0.3$ when attacking normally trained models and $p_d=0.1$ when attacking defense models, and the ensemble number $N=30$ in aggregate gradient.
For feature-level attacks, we choose the same layer, \ie, \emph{Mixed\_5b} for Inc-V3, \emph{Conv3\_3} for Vgg-16, \emph{Conv\_4a} for InRes-V2, and the last layer of second block for Res-152.

\subsection{Comparison of Transferability}
\label{sec:results}

To quantitatively compare the transferability between the proposed \proposed{} and the baselines, we choose Inc-v3, IncRes-v2, Res-152, Vgg-16 as the source model, respectively, and attack the other normally trained models (Table~\ref{tab:compareNor}) and defense models (Table~\ref{tab:compareAdv}). Please note that TIM is not included in Table~\ref{tab:compareNor} because it is designed for defense models. For qualitative comparison, please refer to the supplementary.

\noindent\textbf{Attacking Normally Trained Models.}
We follow the settings in \cite{DBLP:conf/eccv/GaoZSLS20}, \ie, $\beta=10$, $\gamma=16$ for PIM, and $\beta=2.5$, $\gamma=2$ for PIDIM which is the combined version of PIM and DIM. The project kernel size $k_{w}=3$ for the two methods.
As shown in Table~\ref{tab:compareNor}, our method significantly outperforms the other methods for the transferable attack, improving the success rate by 9.5\% on average. 
In particular, the success rate of {\proposed} against each of the normally trained models consistently achieves over 90\% with the source model of Vgg-16, while the other methods may drop to around 60\% in the transferable attack.
Our method can be easily adapted to other methods to further improve the transferability, \eg, \proposed{}+PIDIM is the combination of \proposed{} and PIDIM ($\beta=2.5$, $\gamma=2$ and $k_{w}=3$), which results in 1\% $\sim$ 10\% increase of success rate. Finally, compared to other feature-level attacks, \ie, NRDM and FDA, \proposed{} performs the best in ALL cases (white-box attack and black-box attack), which demonstrates the effectiveness of the proposed aggregate gradient in locating critical object-aware features that dominate the decision across different models.

The results in Table~\ref{tab:compareNor} also give insight into relation between model complexity and transferability, \ie, less complicated/smaller models tend to yield higher transferable adversarial examples (on the premise that the models should achieve similar classification accuracy). For instance, it gets the highest success rate when using Vgg-16 as the source model. Intuitively, those larger models (\eg, IncRes-v2 and Res-152) provide a larger search space, making it more difficult to avoid local optimum.

\begin{table}[t]\small
  \centering
  {\tabcolsep 4pt
  \caption{Success rate of different attacks against defense models when using an ensemble model which contains Res-50, Res-152, Vgg-16, and Vgg-19. The best results are highlighted in bold.}
  \label{tab:ens}
  \renewcommand\arraystretch{0.85}
  \newcommand{\tabincell}[2]{\begin{tabular}{@{}#1@{}}#2\end{tabular}}
  \begin{tabular}{>{\columncolor{mygray}}c|>{\columncolor{mygray}}c|>{\columncolor{mygray}}c|>{\columncolor{mygray}}c|>{\columncolor{mygray}}c|>{\columncolor{mygray}}c}
    \toprule
    \rowcolor{white}
      Attack &  \tabincell{c}{Adv-\\Inc-v3}  & \tabincell{c}{Adv-\\IncRes-v2}  & \tabincell{c}{Ens3-\\Inc-v3} & \tabincell{c}{Ens4-\\Inc-v3} & \tabincell{c}{Ens-\\IncRes-v2} \\
    \midrule
    \rowcolor{white}
     MIM & 71.0\% & 69.6\% & 70.6\% & 71.3\% & 51.7\% \\
     \rowcolor{white}
     DIM & 81.2\% & 83.6\% & 81.6\% & 79.8\% & 65.9\% \\
     \rowcolor{white}
     TIM & 68.0\% & 63.9\% & 70.7\% & 72.7\% & 60.7\%  \\
     \rowcolor{white}
     PIM & 72.1\% & 66.4\% & 76.3\% & 79.3\% & 68.2\% \\
     \rowcolor{white}
     TIDIM & 74.4\% & 68.3\% & 75.9\% & 77.8\% & 67.1\% \\
     \rowcolor{white}
     PITIDIM & 72.1\% & 66.4\% & 76.3\% & 79.3\% & 68.2\% \\
     {\proposed} & \textbf{90.9\%} & \textbf{90.0\%} & \textbf{88.0\%} & \textbf{88.4\%} & \textbf{75.8\%} \\
    \bottomrule
  \end{tabular}}
  \vspace{-4mm}
\end{table}

\noindent\textbf{Attacking Defense Models}
Since defense models are trained adversarially thus showing strong robustness to adversarial examples. In Table~\ref{tab:compareAdv}, the other feature-level attacks are not listed because of their bad performance. In PIM, we follow the suggestion from \cite{DBLP:conf/eccv/GaoZSLS20} to remove the momentum term since it may hinder the performance of attacking defense models. The settings of PIM and its combinations are $\beta=10$, $\gamma=16$, and $k_{w}=7$.
As shown in Table~\ref{tab:compareAdv}, {\proposed} or the correspondingly combined version \proposed{}+PITIDIM ($\beta=2.5$, $\gamma=2.0$, and $k_{w}=7$) outperforms other methods. In most cases, {\proposed}+PITIDIM and \proposed{} rank the top two, and {\proposed}+PITIDIM would further improve the transferability as compared to \proposed{}.
In average, our approach increases the success rate by about 12.8\% in attacking defense models as compared to the other methods.

Since the relatively lower transferability in attacking defense models, we further improve transferability by generating adversarial examples on an ensemble of models~\cite{DBLP:conf/iclr/LiuCLS17}, which prevents adversarial examples from falling into local optimum of a single model.
Following the setting in \cite{DBLP:conf/eccv/GaoZSLS20}, \ie, $\beta=5$ and $\gamma = 8$ for PIM and its combined versions, the results are shown in Table~\ref{tab:ens}, where all methods are improved, and our method still outperforms the others.

\begin{figure}[]
    \setlength{\abovecaptionskip}{0.2cm}
    \setlength{\belowcaptionskip}{-0.cm}
	\centering
	\subfigure{
		\includegraphics[width=0.22\textwidth]{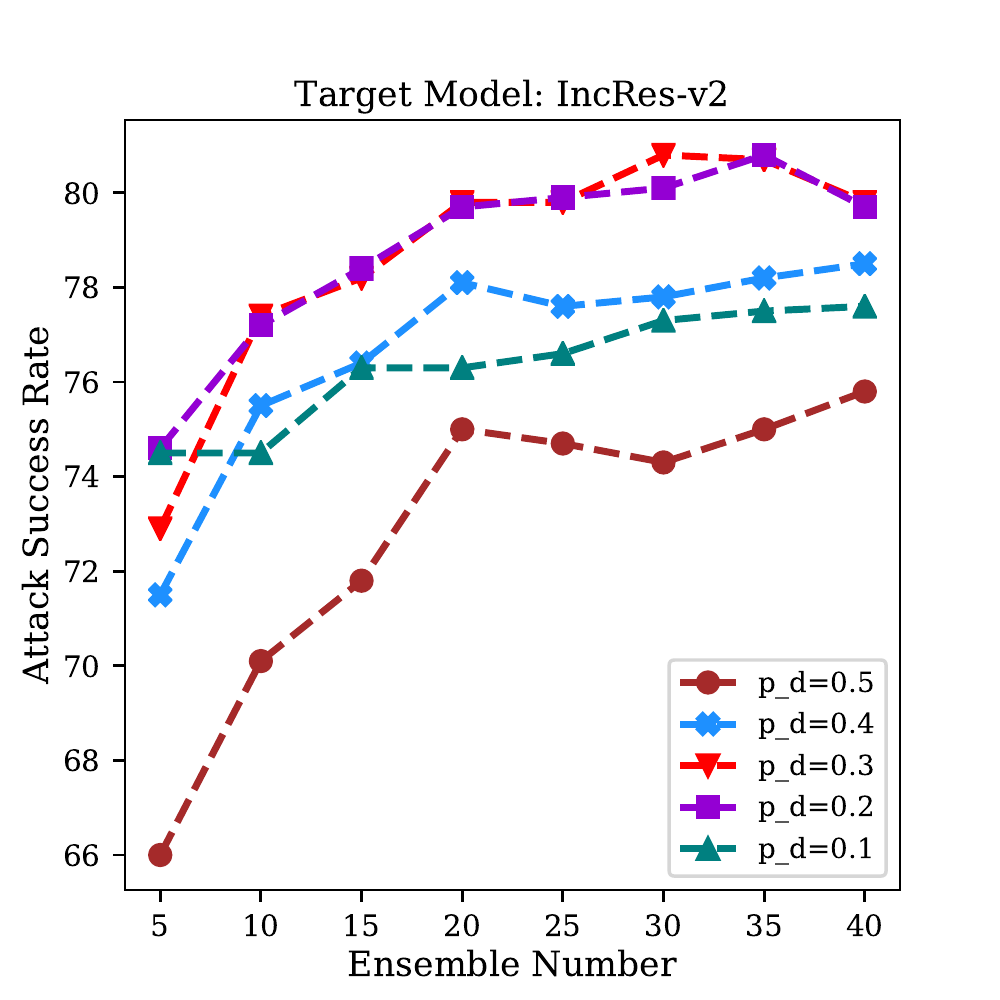}
			\label{fig:para1}
	}
 	\hspace{-3mm}
	\subfigure{
	\includegraphics[width=0.22\textwidth]{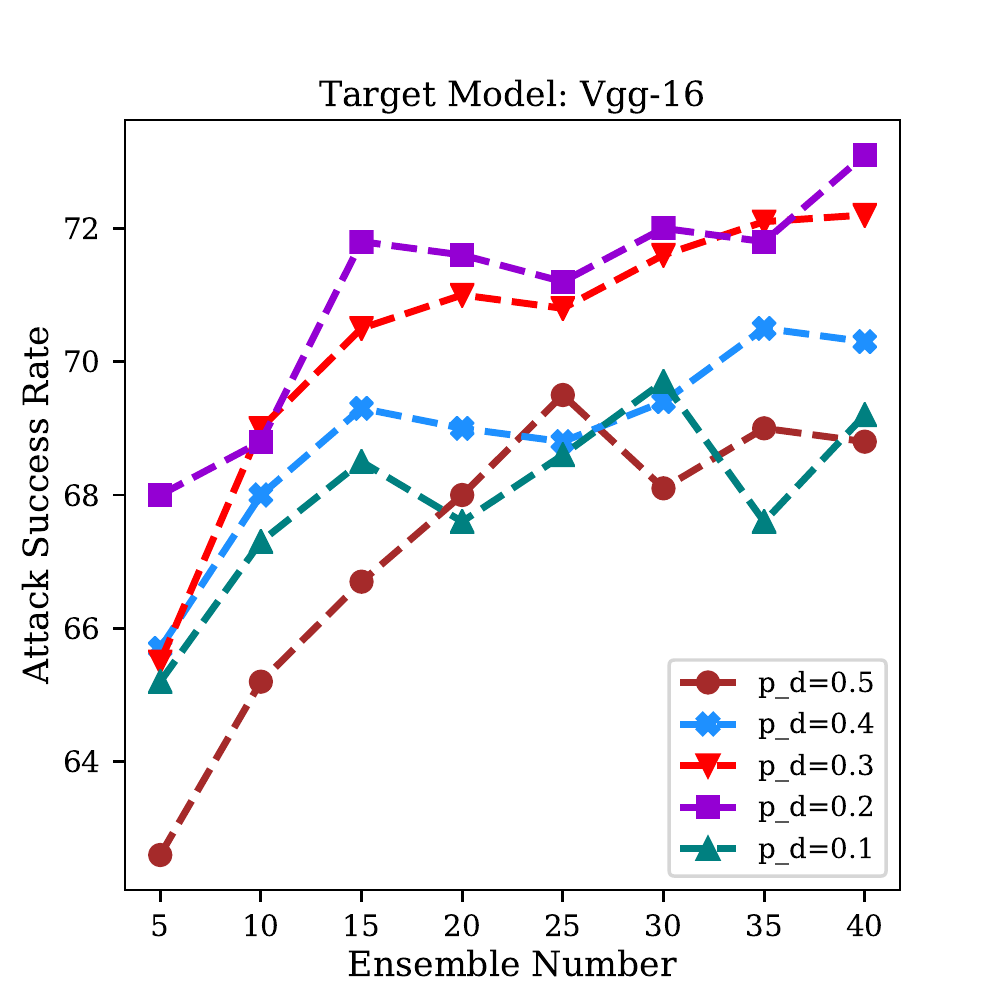}
			\label{fig:para2}
	}
	\subfigure{
	\includegraphics[width=0.22\textwidth]{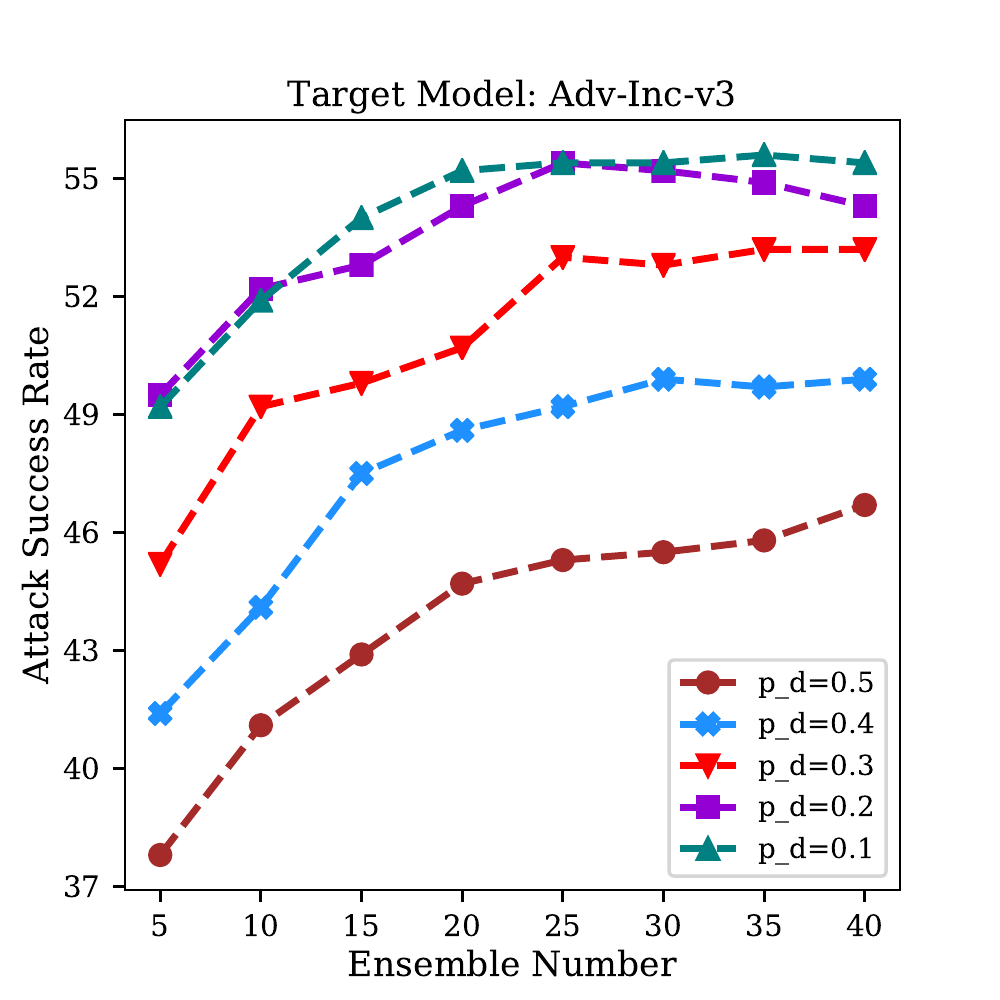}
			\label{fig:para3}
	}
 	\hspace{-3mm}
	\subfigure{
	\includegraphics[width=0.22\textwidth]{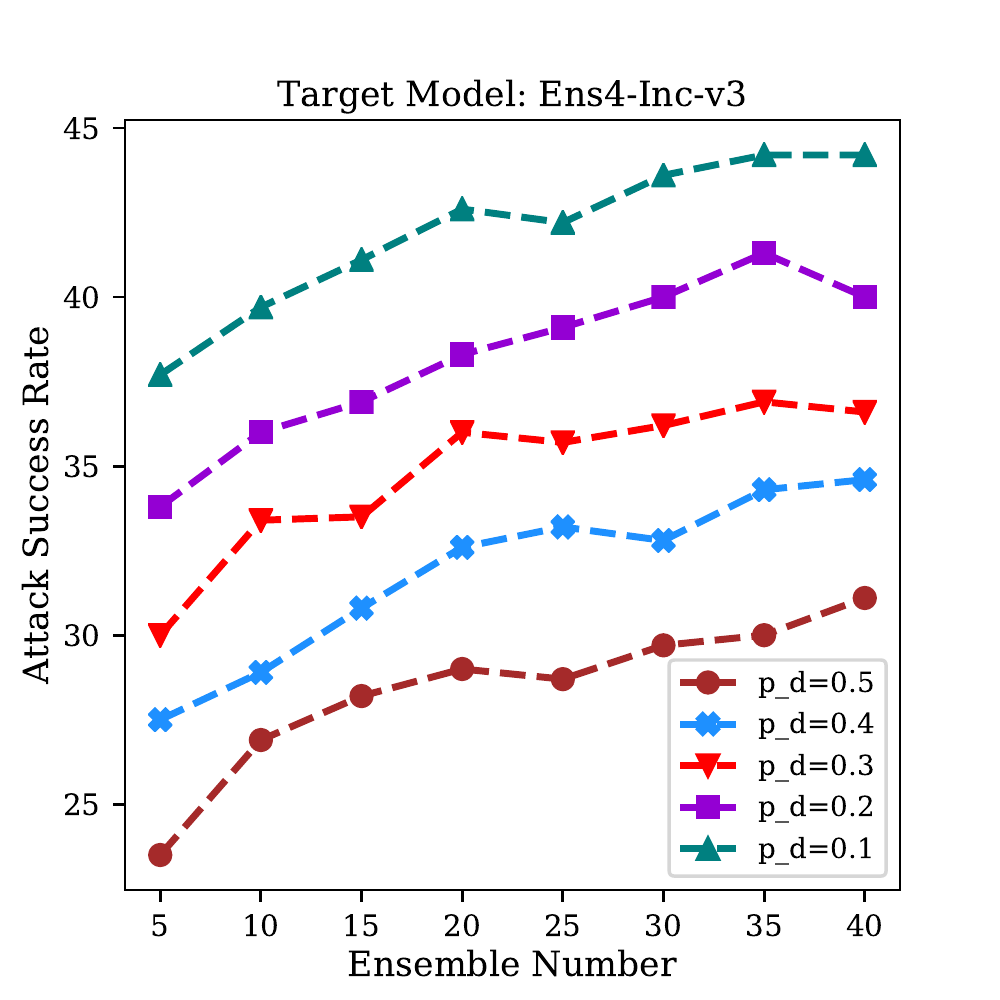}
			\label{fig:para4}
	}
    \caption{Effect of the drop probability and ensemble number on attack success rate. The adversarial examples are generated by {\proposed} with different parameter setting against the source model Inc-v3. The drop probability changes from 0.1 to 0.5 with the step of 0.1, and the ensemble number varies from 5 to 40 with the step of 5. The top row are success rates of attacking two normally trained models IncRes-v2 and Vgg-16, and the bottom are results of attacking two defense models Adv-Inc-v3 and Ens4-Inc-v3. }
    \label{fig:para}
    \vspace{-5mm}
\end{figure}

\subsection{Effect of Parameters in Aggregate Gradient}\label{sec:hyper}
There are three parameters, \ie, drop probability $p_d$, ensemble number $N$, and layer $k$ (Eq.~\ref{equ:agg_grad}), which affects the performance of the proposed \proposed{}. For the first two parameters, we adopt Inc-v3 as the source model and modify $p_d$ from 0.1 to 0.5 with the step of 0.1. For each $p_d$, $N$ is iterated from 5 to 40 with the step of 5. Fig.~\ref{fig:para} illustrates the effect of $p_d$ and $N$ by attacking IncRes-v2, Vgg-16, Adv-Inc-V3, and Ens4-Inc-v3, where the drop probability and ensemble number affect the success rate in a roughly consistent manner across different target models.

More specifically, the drop probability $p_d$ plays an important role in affecting the success rate, and such effect tends to be consistent across different target models. A large $p_d$ (\eg, 0.5) will destroy important structural information of an image, thus significantly reducing the success rate. Therefore, an optimal $p_d$ for attacking normally trained models is between 0.2 and 0.3, and it should be around 0.1 if attacking defense models.
For the ensemble number $N$, a larger $N$ tends to yield a higher success rate but will saturate gradually.
Finally, we determine the ensemble number $N=30$, drop probability $p_d=0.3$ against normally trained models, and  $p_d=0.1$ against defense models.

\begin{figure}[t]
	\centering
	\subfigure{ \includegraphics[width=0.22\textwidth]{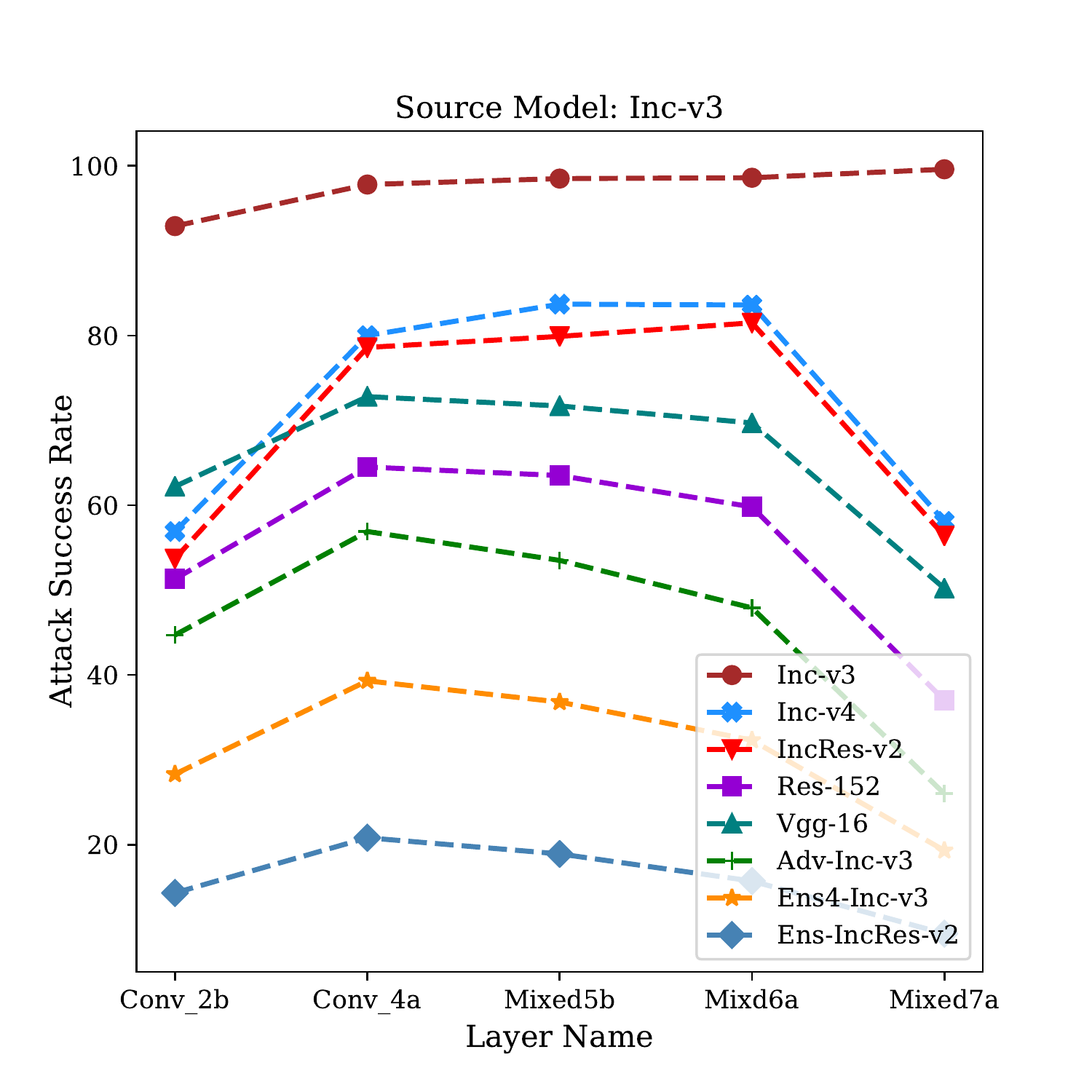}
			\label{fig:ablation_inc}
	}
	\hspace{-3mm}
	\subfigure{ \includegraphics[width=0.22\textwidth]{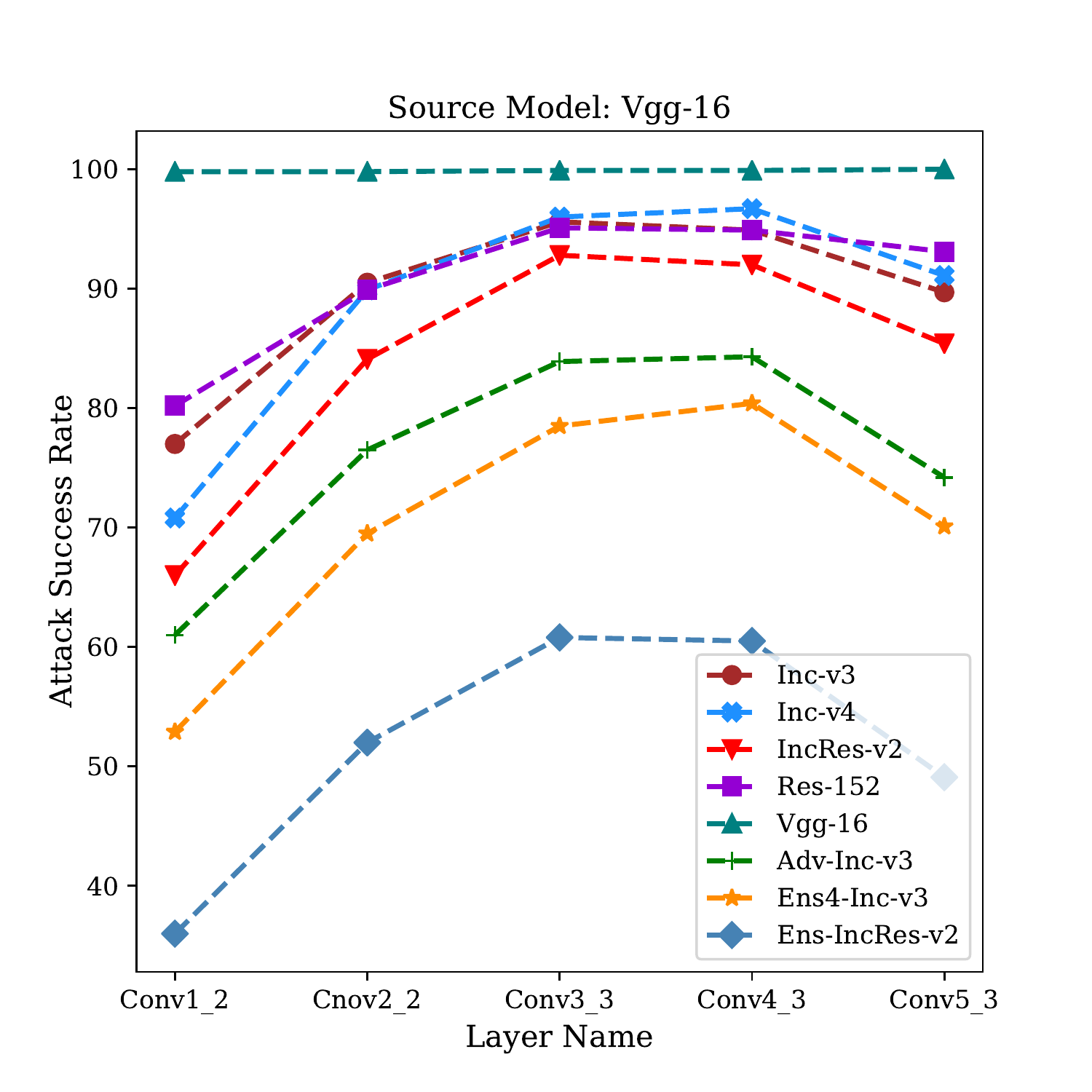}
			\label{fig:ablation_vgg}
	}	
    \caption{Effect of layer choice on attack success rate. Different layers from the source models (\ie, Inc-v3 and Vgg-16) are selected to generate adversarial examples, whose success rates are reported against different target models.
    }
    \vspace{-5mm}
    \label{fig:layers}
\end{figure}

Feature-level attacks are significantly affected by the choice of feature layer $k$ since early layers of DNNs may be working to construct a basic feature set which is usually data-specific, and further layers may process these extracted features to maximize the classification accuracy of the model which makes the features become model-specific~\cite{DBLP:conf/iclr/InkawhichLCC20}. Therefore, early layers have not learned salient features and semantic concepts of the true classes, and later layers are model-specific that should be avoided in transferable attacks. By contrast, middle layers have well-separated class representations and they are not highly correlated to the model architecture, thus middle layers are the best choice to be attacked for better transferability. The same conclusion can be drawn from Fig.~\ref{fig:layers}, which reports the success rate of attacking different target models by adversarial examples optimized on different source model layers. Based on this conclusion, we first select a few middle layers for each source model and determine the final attacked layer according to the empirical results.

\begin{figure}[]
	\centering
	
	\subfigure{ \includegraphics[width=0.38\textwidth]{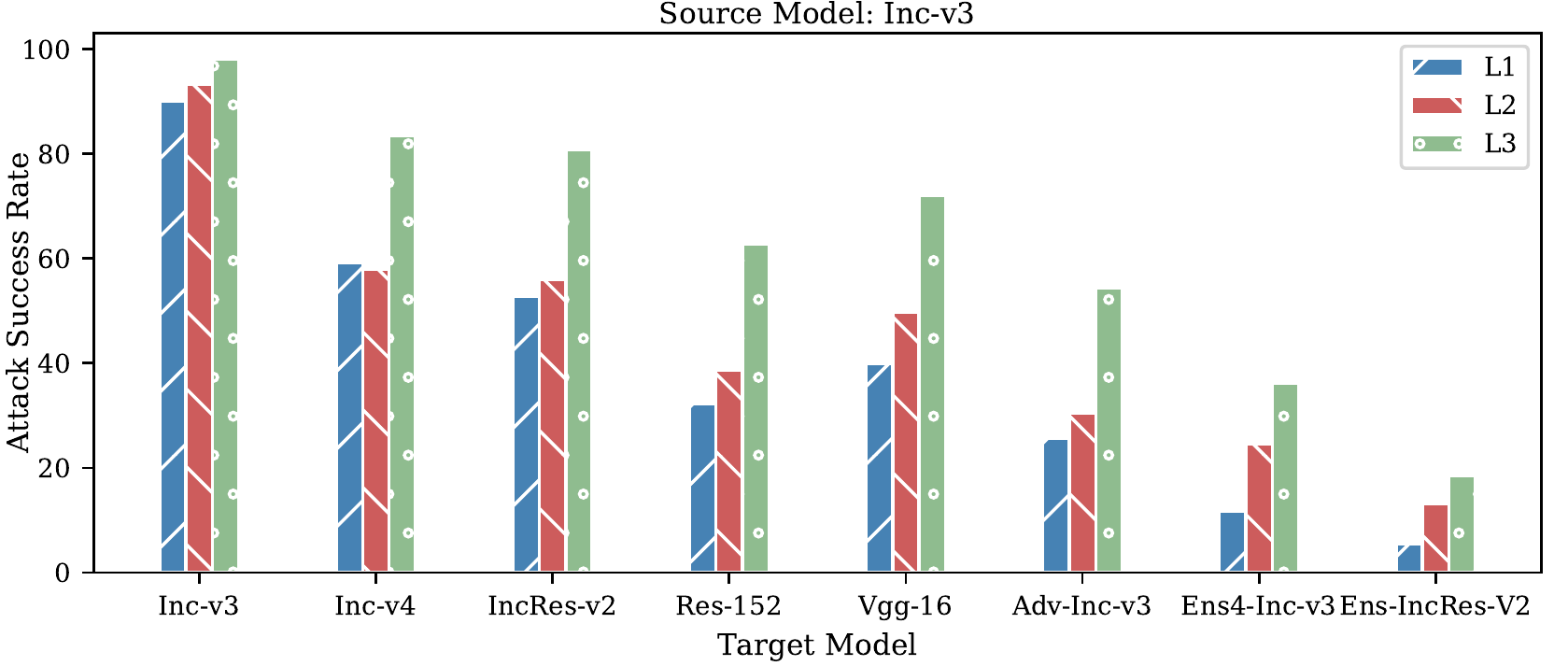}
			\label{fig:ablation_inc}
			\vspace{-3cm}
	}

	\subfigure{ \includegraphics[width=0.38\textwidth]{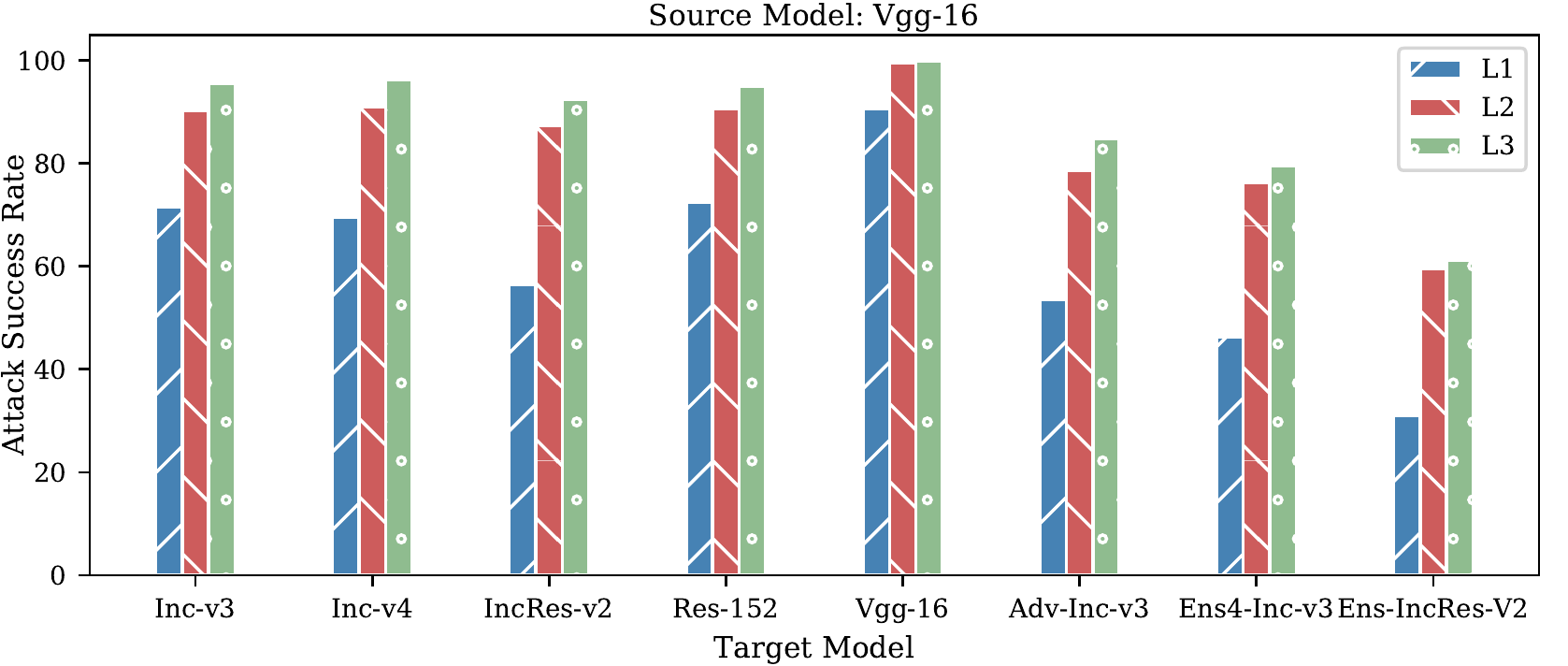}
			\label{fig:ablation_vgg}
	}	
    \caption{Effect of aggregate gradient on attack success rate. $\mathcal{L}_1$ optimizes feature distortion without gradient guidance, $\mathcal{L}_2$ uses raw gradient, and $\mathcal{L}_3$ adopts aggregate gradient.}
    \vspace{-5mm}
    \label{fig:ablation}
\end{figure}

\subsection{Ablation Study}\label{sec:abl}

The key of the proposed \proposed{} is the aggregate gradient $\Delta$, which significantly boosts the transferability as demonstrated in the aforementioned results. To highlight the contribution of aggregate gradient, we conduct the ablation study to compare the performance of objectives with and without aggregate gradient. We construct three objective functions as express in the following, where $\mathcal{L}_1$ optimizes the feature divergence without constraints like most of the baseline methods, and $\mathcal{L}_3$ is equivalent to our proposed loss (Eq.~\ref{equ:loss}). The $\mathcal{L}_2$ uses non-aggregate gradient $\Delta_{clean}$, \ie, gradient from the original clean image. Fig.~\ref{fig:ablation} shows the success rate using the three losses, respectively.
\begin{align}
\label{equ:abs_loss} \mathcal{L}_{1} & = \sum \left|f_{k}(x)- f_{k}(x^{adv})\right|, \\
\label{equ:grad_loss} \mathcal{L}_{2} & = \sum ( \Delta_{clean} \odot (f_{k}(x)- f_{k}(x^{adv}))), \\
\label{equ:ens_loss} \mathcal{L}_{3} & = \sum ( \Delta \odot (f_{k}(x)- f_{k}(x^{adv}))).
\end{align}

The proposed loss $\mathcal{L}_{3}$ outperforms the others by a large margin in all case, indicating effectiveness of the proposed aggregate gradient.

\section{Conclusion}\label{sec:conclusion}
In this work, we proposed a Feature Importance-aware Attack ({\proposed}) to generate highly transferable adversarial examples, whose effectiveness are demonstrated in our exhaustive experiments.
The proposed {\proposed} explores the feature importance through aggregate gradient across various of classification models and introduces such transferable information into the search of adversarial examples.
Consequently, the optimization process is guided towards disrupting the critical object-aware features that dominate the decision of the models, thus gaining remarkable transferability.
We conducted extensive experiments to demonstrate the superior performance of \proposed{} as compared to those state-of-the-art methods, and our method can serve as a benchmark for evaluating the robustness of various models.

\section*{Acknowledgments}
This work was supported by National Natural Science of China (Grants No. 62122066, U20A20182, 61872274, U20A20178, 62032021, and 62072395), National Key R\&D Program of China (Grant No. 2020AAA0107705), the Fundamental Research Funds for the Central Universities (Grant No. 2021QNA5016).

%

{\small
\bibliographystyle{ieee_fullname}
\bibliography{egbib}
}

\end{document}